\documentclass[
twocolumn,
]{ceurart}

\usepackage{listings}
\usepackage{graphicx}
\usepackage{expex}
\usepackage{subcaption}
\usepackage{gb4e}

\begin{document}

\copyrightyear{2024}
\copyrightclause{Copyright for this paper by its authors.
  Use permitted under Creative Commons License Attribution 4.0
  International (CC BY 4.0).}
\conference{CLiC-it 2024: Tenth Italian Conference on Computational Linguistics, Dec 04 — 06, 2024, Pisa, Italy}

\title{Did somebody say `Gest-IT'? \\ A pilot exploration of multimodal data management}
\author[1]{Ludovica Pannitto}[%
orcid=0000-0002-5760-6447,
email= ludovica.pannitto@unibo.it,]
\cormark[1]
\author[1]{Lorenzo Albanesi}[%
email= lorenzo.albanesi@studio.unibo.it,
]
\author[1]{Laura Marion}[%
email= laura.marion@studio.unibo.it,
]
\author[1]{Federica Maria Martines}[%
email= federica.martines2@studio.unibo.it,
]
\author[1]{Carmelo Caruso} [%
email= carmelo.caruso@unibo.it
]
\author[2]{Claudia S. Bianchini} [%
orcid= 0000-0002-4783-1202, 
email= claudia.savina.bianchini@univ-poitiers.fr
]
\author[1]{Francesca Masini}[%
orcid=0000-0002-2354-4612,
email= francesca.masini@unibo.it,
]
\author[1]{Caterina Mauri}[%
orcid=0000-0002-8226-7846,
email= caterina.mauri@unibo.it,
]

\address[1]{Alma Mater Studiorum - University of Bologna}
\address[2]{University of Poitiers, FoReLLIS Laboratory}


\cortext[1]{Corresponding author.}

\begin{abstract}
The paper presents a pilot exploration of the construction, management and analysis of a multimodal corpus. Through a three-layer annotation that provides orthographic, prosodic, and gestural transcriptions, the \textit{Gest-IT} resource allows to investigate the variation of gesture-making patterns in conversations between sighted people and people with visual impairment. After discussing the transcription methods and technical procedures employed in our study, we propose a unified CoNLL-U corpus and indicate our future steps.
\end{abstract}


\maketitle

\section{Introduction}
\label{sec:intro}

Corpora represent the main tool for linguists to observe language in its real use and verify its general trends on both a quantitative and qualitative basis~\cite{ludeling2009corpus}. Today, written language corpora are the most used, thanks to the greater availability of written data and the ease of processing. However, in speech, speakers appeal to numerous semiotic sources (e.g., spoken channel, gestures, proxemics, facial expressions, etc.) to create and convey meaning, and written corpora fail to account for this richness of modalities.
To effectively study how language works, one should observe these different semiotic sources independently of each other and take their interactions into account~\cite{bezemer2018multimodality}. To capture this complexity, it is necessary to go beyond written data and use multimodal corpora, namely collections of audio-visual linguistic data that allow to both hear and see linguistic productions.

Multimodal corpora can be used to analyze a wide variety of linguistic phenomena, especially those related to the use of body in human communication and interaction, and to the way bodily communication and spoken language interact to generate meaning. They are, therefore, the primary sources for the analysis of co-speech gestures~\cite{abner2015gesture} and Sign Languages.
Following~\cite{abuczki2013overview,foster2007corpus}, we define a multimodal corpus as `an annotated collection of coordinated content on communication channels including speech, gaze, hand gesture and body language'. Multimodal corpora come in various shapes, depending on the nature of the communication channel captured by the resource. For the sake of this article, we specifically restrict our attention to resources including both a video and audio recording of linguistic content\footnote{There exists a wide variety of multimodal resources for spoken and signed language, many of them openly available to the community through initiatives such as CLARIN-ERIC (\url{https://www.clarin.eu/}). For a collection of available multimodal resources see \url{https://www.clarin.eu/resource-families/
multimodal-corpora} (spoken language) and \url{https://www.clarin.eu/resource-families/sign-language-resources} (sign language), while a list of audio-only resources of spoken language can be found at \url{https://www.clarin.eu/resource-families/spoken-corpora}}. The long tradition of analyzing written and spoken data has, over time, led to the development of transcription systems (such as \textsc{IPA}, orthography, and various prosodic conventions), which have become recognized standards within the linguists' community. These systems allow for an ``objective'' description of speech, independently of any considerations on the functions or the meanings of the described elements. For gestural data, some transcription systems have also been proposed, such as the Linguistic Annotation System for Gestures (LASG; \cite{Bressem2013}), but none of them has attained enough acceptance to qualify as a standard~\cite{abuczki2013overview}. In the absence of an effective transcription system, hybrid solutions are often employed, situated between transcription and annotation, where the choices for describing gestural forms reflect their attributed functions (e.g.: a ``shrug'' is a shoulder movement, but this label is often used to refer to any pragmatic gesture of epistemic denial performed by moving the shoulders, without specifying either the characteristics of the shoulder movement or the movements of other body parts that may have contributed to the execution of the gesture). Similar challenges arise in studies on Sign Languages. Although there is a larger number of transcription systems for the latter (see, for instance, the review in~\cite{bianchini:hal-04602726}), none of them has achieved the status of a universal standard. Additionally, attempts to adapt these systems to the transcription of gestures have so far been limited and not particularly successful. The lack of a transcription standard for gestures, that describes them independently of their function or meaning, hinders the ability to precisely investigate the relationship between speech and gesture.

Another aspect concerns the nature of language data captured in multimodal resources: as the collection and standardization process for this kind of linguistic data is, by its very nature, much more complex, resources are often tailored to specific purposes and therefore involve task-oriented interactions (e.g., describing objects as in the \textit{NM-MoCap-Corpus}~\cite{freigang2014your}; spatial comunication tasks as in the \textit{SaGA Corpus}~\cite{lucking2010bielefeld}), thus capturing interactions that may be naturalistic but are inherently non-ecological, i.e. not naturally-occurring \cite{dubois2023typology,troiani2023representing}. Often, participants are asked to wear special devices such as headsets or trackers during the recordings~\cite{bertrand2008cid,knight2009nottingham}, clearly altering the spontaneity of the interaction.

The aim of the \textit{Gest-IT} project is to build a multimodal corpus of ecological data, allowing for the integrated analysis of verbal and gestural communication in spontaneous interactions. In this paper, we will focus on the protocol of multimodal data management that we tested for this resource. We will first discuss the main existing multimodal resources (Section~\ref{sec:related}), showing how, as of today, there doesn't seem to be any ecological, accessible, multimodal corpus for Italian. We will then introduce the \textit{Gest-IT} pilot resource and present its main features with respect to existing resources (Section~\ref{sec:gestit}). Section~\ref{sec:schema} outlines the main design choices taken for the creation of our resource and Section~\ref{sec:future} describes the path ahead.

\section{Multimodal resources: problems and overview}
\label{sec:related}

Multimodal corpus research faces two major problems: (i) the lack of existing transcription and annotation standards (tools, formats and schemes), especially for coding nonverbal behavior~\cite{abuczki2013overview}; and (ii) the time consuming nature of transcription and annotation process, which is responsible for the relatively small sizes of searchable multimodal corpora that are currently available.

Specifically with respect to point (i), a major problem concerning available resources is the non-separation between the identification and description of gestures on the one hand, and their interpretation on the other. Indeed, in many resources and studies a particular gestural pattern is transcribed based on its \textit{function}, i.e. its interpretation, rather than on a description of the `objective' aspects that characterize its `form'. However, if we aim to provide an integrated analysis of verbal and nonverbal communication, it is crucial that – just as we employ \textsc{ipa} or simplified orthographic transcriptions for verbal signs – we establish a standard to transcribe nonverbal signs in order to then annotate and interpret them. 
Furthermore, in most resources gesture is transcribed only with reference to verbal behaviour: the very identification of the gestures depends on their association, according to the annotator's subjective filtering, to an identifiable verbal sequence.  In the \textit{PoliModal corpus}\footnote{\url{https://github.com/dhfbk/InMezzoraDataset}}~\cite{trotta2020adding}, a resource including transcripts of 14 hours of TV face-to-face interviews from the Italian political talk show \textit{Mezz'ora in più}, for instance, gestures are annotated if they are judged as having a communicative intention~\cite{allwood2001capturing} (displayed or signalled), or a noticeable effect on the recipient. Once a gesture has been selected, it is annotated with functional values, as well as features that describe its behavioural shape and dynamics. 
The descriptions provided for gesture annotation, moreover, seem to be an approximation of the movement: gestures are often described relying on the annotator's categorization and not using meaningful and objective parameters. 
For example, in the MUMIN coding~\cite{allwood2007mumin} scheme used in the \textit{PoliModal Corpus} and reported in Table~\ref{tab:mumin}, a number of possible values for each behaviour attribute are defined, but these fail to describe the entire range of possibilities (i.e., only three values are provided for face movements) or excessively simplify the description (i.e., the value \textit{complex} is used to capture movements where several trajectories are combined, thus leaving unspecified whether they combine sequentially or in a non-linear trajectory for instance).
Similar code schemes are used in the \textit{Corpus d'interactions dialogales} (CID,~\cite{bertrand2008cid}) and in the \textit{Hungarian Multimodal Corpus}~\cite{papay2011hucomtech}.

For resources such as \textit{Natural Media Motion-Capture Corpus} (NM-MoCap-Corpus~\cite{freigang2014your}), \textit{Bielefeld Speech and Gesture Alignment Corpus} (SaGA~\cite{lucking2010bielefeld}) and \textit{BAS SmartKom Public Video and Gesture corpus} (SKP~\cite{schiel2002smartkom}), researchers decided to adopt McNeill's categories~\cite{mcneill2013gesture} or a schema inspired by them~\cite{mlakar2012capturing,mlakar2019towards}. 
In addition, some of the Swedish data in the \textit{Thai/Swedish child data corpus}~\cite{fivser2020overview} were partially annotated thanks to the standard notation CHAT~\cite{macwhinney1998computational}. In the CORMIP~\cite{lo2021prosody} resource, instead, each gesture is segmented according to gesture phrases and gesture units~\cite{kendon1980gesticulation}. Gestures are then classified solely based on iconicity, classifying them as `Pictorial', `Non-Pictorial' or `Conventional'. While they claim to avoid categorization of gesture functions or conventionality, the description of their lables (see Table~\ref{tab:cormip}) seems to contradict this statement~\cite{lo2022corpus}.
Lastly, as far as Italian is concerned, the \textit{Padova Multimodal Corpus}~\cite{AckerleyCoccetta2007,coccetta2011multimodal} has to be mentioned, where textual transcriptions are enriched with annotations about a number of non-verbal components, there including also aspects such as gaze and gestures.
The MultiModal MultiDimensional (M3D) labelling scheme\footnote{\url{https://osf.io/ankdx/}}~\cite{rohrer2020multimodal,rohrer2023visualizing} tries to decouple gesture transcription in the three different dimensions of its form, its relation to spoken prosody and its semantic or pragmatic functions. As reported in their manual, however, the transcriber is required to make choices, on the \textit{form} layer, such as which is the predominant articulator (e.g., left or right hand, or both) or to choose for the articulator one of the provided forms, one of which is labeled as \textit{iconic OK shape}.

\begin{table}[]
\caption{MUNIN~\cite{allwood2007mumin} coding scheme}
\scriptsize{
\begin{tabular}{p{0.25\linewidth}p{0.6\linewidth}}
\toprule
Behaviour attribute & Behaviour value \\
\midrule
General face & Smile, Laugh, Scowl, Other \\
Eyebrow movement & Frown, Raise, Other \\
Eye movement & Extra-Open, Close-Both, Close-One, Close-Repeated, Other \\
Gaze direction & Towards-Interlocutor, Up, Down, Sideways, Other \\
Mouth openness & Open mouth, Closed mouth \\
Lip position & Corners up, Corners down, Protruded, Retracted \\
Head movement & Down, Down-Repeated, BackUp, BackUpRepeated, BackUp-Slow, Forward, Back, Side-Tilt, Side-TiltRepeated, Side-Turn, Side-Turn-Repeated, Waggle, Other \\
Handedness & Both hands, Single hands \\
Hand movement trajectory & Up, Down, Sideways, Complex, Other \\
Body posture & Towards-Interlocutor, Up, Down, Sideways, Other \\
\bottomrule
\end{tabular}%
}
\label{tab:mumin}
\end{table}

\begin{table}[]
    \centering
    \caption{Gestures classification in CORMIP \cite{lo2022corpus}}
    \scriptsize{
    \begin{tabular}{p{0.3\linewidth}p{0.6\linewidth}}
    \toprule
        \textit{Pictorial} & image-like shapes, or boundaries of a real-world object or action.\\
        \textit{Non-Pictorial} & rythmic movements (i.e., batonic) or geometric forms. Deictic gestures also fall within this category.\\
        \textit{Conventional} & gestures with a degree of conventionality that allows to associate, in a specific linguistic system, a semantic value to tehm (e.g., the `okay' sign).\\
        \bottomrule
    \end{tabular}
    }
    \label{tab:cormip}
\end{table}

The challenge of transcription becomes even more significant when dealing with multimodal corpora representing sign language. Typically, this issue is addressed using glosses, a form of sign-to-word translation that provides information about the meaning of signs without indicating their form~\cite{bianchini:hal-04602726}. However, over the years, some systems have been developed to represent the shape of signs. Most of these systems focus primarily on the hands~\cite{chevrefils2021body}, which are only a small part of the articulators contributing to meaning. Among these systems, Typannot~\cite{boutet2018systemes} stands out as it offers a comprehensive description of the entire set of body parts— from fingers to toes, including the head and torso — used to transcribe both sign languages and co-verbal gestures.

\section{Towards the \textit{Gest-IT} corpus: blind and sighted speakers}
\label{sec:gestit}

We aim at building a corpus consisting of maximally ecological interactions, transcribed on three separate layers aligned to each other: (i) an orthographic transcription; (ii) a prosodic transcription, and (iii) a gestural transcription. At present, we are still in an initial, exploratory phase, but we already addressed the most important decisions to be made.

The first decision concerned the informants to be recorded. In order to be able to investigate whether the ability to see and the perception of being seen during a communicative exchange can influence gesture production, we decided to take into consideration both sighted and visually impaired L1 speakers in dialogical situations. Gesture is indeed closely linked not only to intersubjective needs, connected to clarity, efficiency and attention-getting functions, but also to cognitive needs: speakers recur to gestures both when the interlocutor is not visible \cite{Alibali2001} and when the speaker is visually impaired \cite{Iverson1998}, thus independently of the interlocutors' ability to see and interpret them. Yet, the actual relation and reciprocal influence between gestures and the perception of being seen has received little attention so far.

We included in the study 6 blind and 8 sighted participants, recruited on a voluntary basis and through a protocol that has been evaluated as compliant with GDPR and ethical requirements\footnote{Positive evaluation of the Bioethics Committee of the University of Bologna n. 0020349, 24/01/2024.}.
The blind group included speakers who were born blind, who acquired blindness later and who are partially-sighted.
The total average age of the participants is mean $= 39$ years old (sd $= \pm18.7$). The average age of the PG is mean $= 55.8$ years (sd $= \pm18$), while the control group has an average age of mean $= \pm26$ years old (sd $= \pm3.9$).
The total gender distribution is $85.7\%$ F and $14.2\%$ M. 
In the blind goup (BG) $100\%$ of the participants are F. In the sighted group (SG) $75\%$ are F and $25\%$ are M.
The total average educational level distribution shows that $64.2\%$ of participants has a bachelor's degree, while $35.7\%$ has a high school diploma. In the BG $83.3\%$ of participants has a high school diploma, while $16.7\%$ of the participants has a bachelor's degree.
In the SG $100\%$ of participants in the control group has a bachelor's degree.

All participants were paired and later involved in 30-minutes seated conversations, to elicit samples of spontaneous speech. As the participants to each dialogue were unlikely to know each other, in order to avoid moments of silence, some questions were prepared to enhance spontaneous conversations (see Appendix~\ref{app:prompts}). Interestingly, speakers recurred to these prompts only in few cases: the interactions developed very spontaneously despite the absence of previous contacts among the interlocutors. 

We built the pairs and the interactional setting according to two parameters:
\begin{itemize}
    \item speakers could belong to the same category of participant (both blind, both sighted) or different categories. We coded these two situations as \textsc{S} (\textit{same}, blind-blind or sighted-sighted conversation) or \textsc{D} (\textit{different}, blind-sighted conversation);
    \item speakers could be facing each other or be seated back-to-back, to ensure that participants could not perceive the other's nonverbal communication. We coded these two situations as \textsc{M} (\textit{masked}, back-to-back situation) or \textsc{U} (\textit{unmasked}, facing situation).
\end{itemize}

\begin{figure}
    \centering  
    \includegraphics[width=0.5\linewidth]{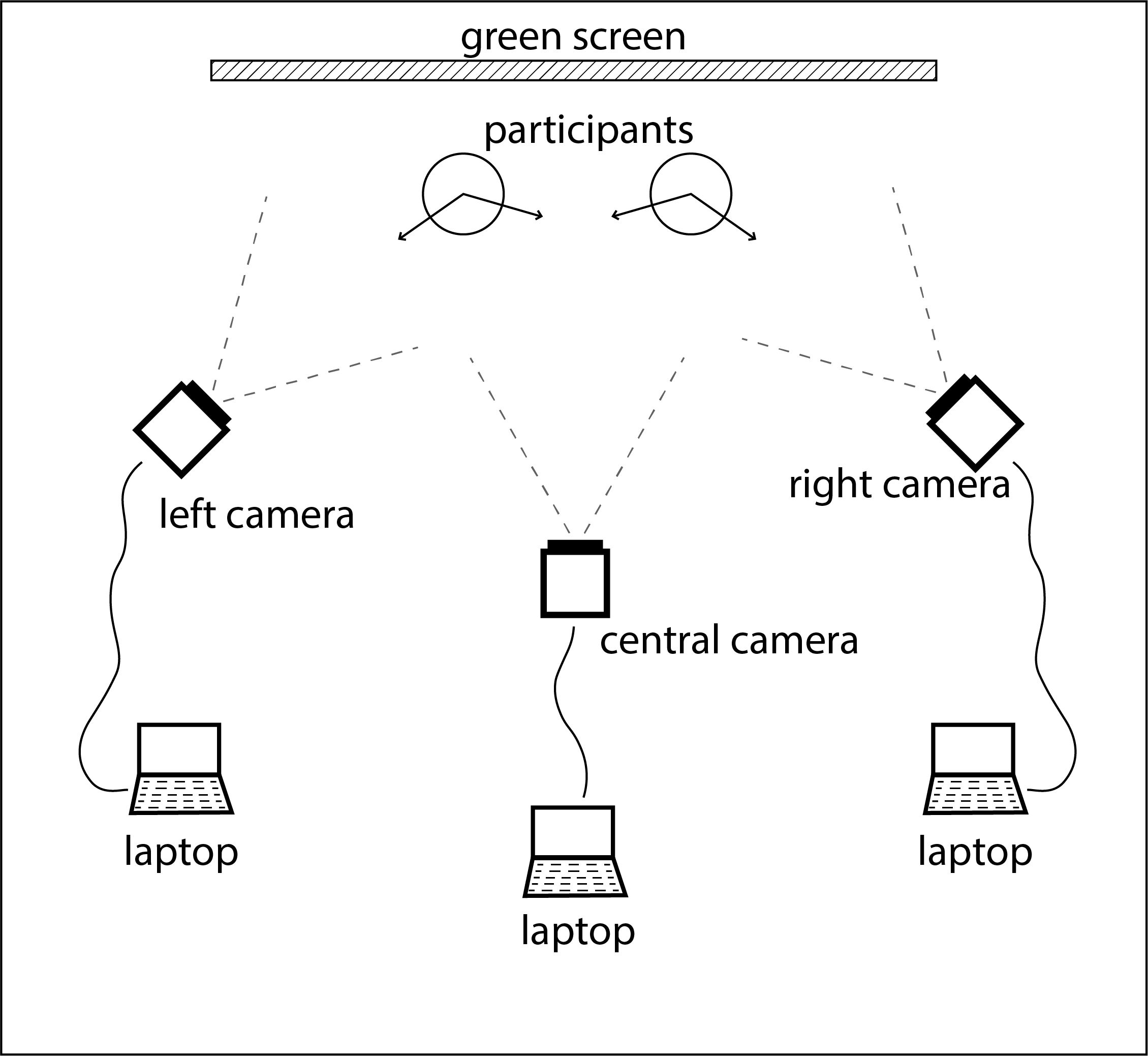}
    \caption {Room setting for the recording sessions, Logitech Brio Stream Webcam were employed for recordings}
    \label{fig:roomsetting}
\end{figure}

\begin{figure}
     \centering
     \begin{subfigure}[b]{0.45\linewidth}
         \centering
         \includegraphics[width=\linewidth]{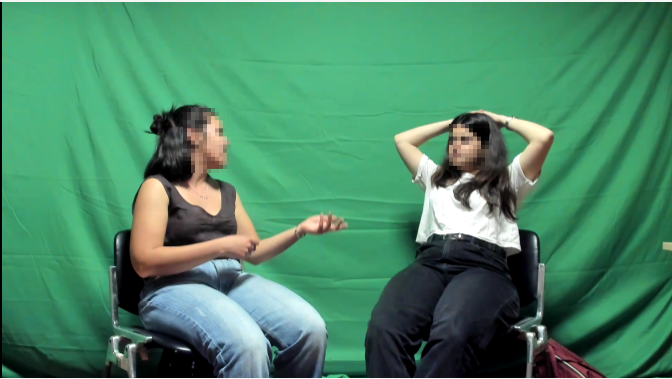}
         \caption{Unmasked scenario, Same sight conditions}
         \label{fig:a}
     \end{subfigure}
     \hfill
     \begin{subfigure}[b]{0.45\linewidth}
         \centering
         \includegraphics[width=\linewidth]{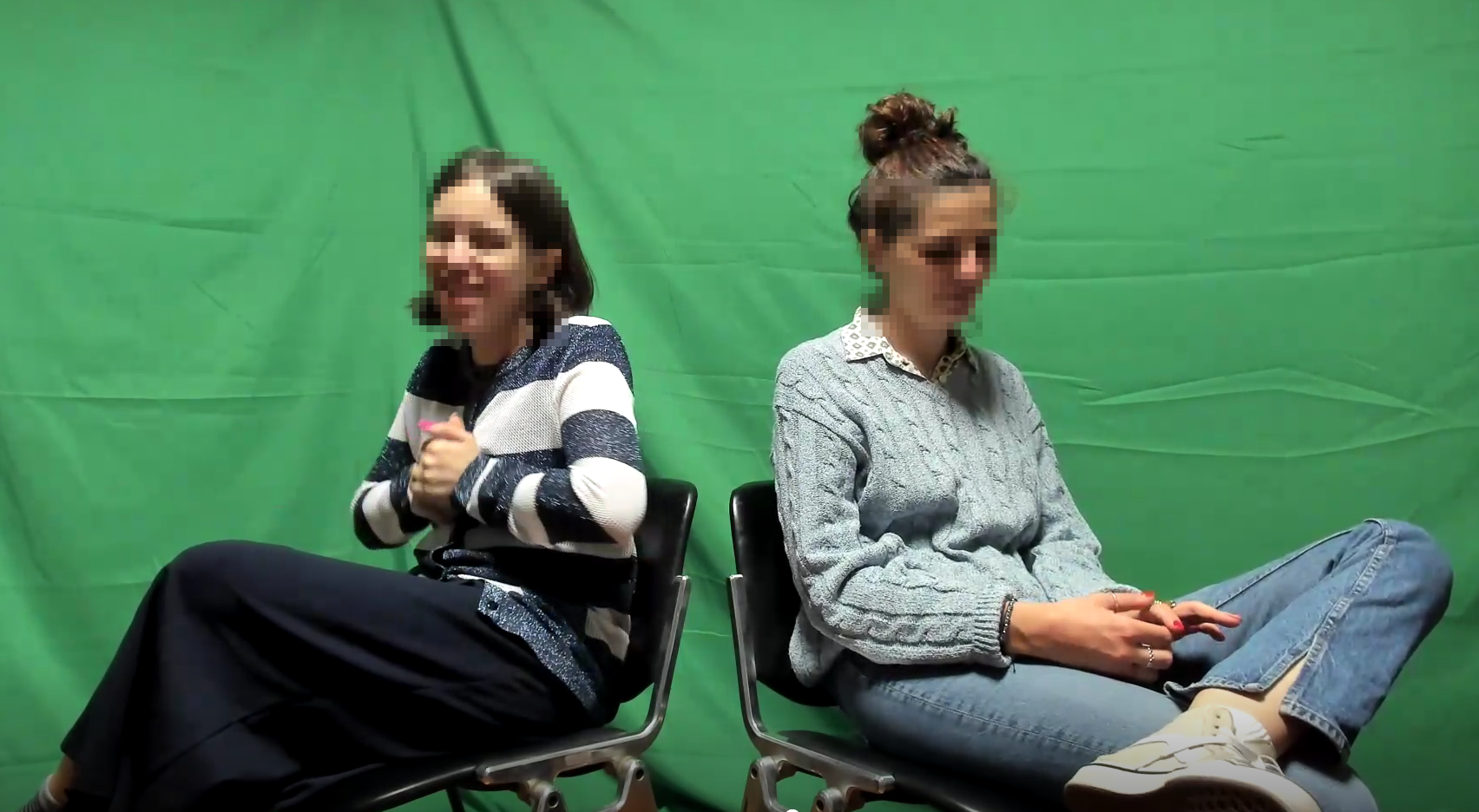}
         \caption{Masked scenario, Different sight conditions}
         \label{fig:b}
     \end{subfigure}
        \caption{Recording scenarios: unmasked and masked, same and different sight conditions}
        \label{fig:riprese}
\end{figure}

We recorded 13 conversations, for a total of roughly 7 hours ($428.15$ minutes), from three points of view: the central camera faced the couple, whereas the other two recorded the left side and the right side (the left and right cameras were located so that they could capture the participants frontally, see Figures~\ref{fig:roomsetting} and~\ref{fig:riprese}).
The goal was to take the participants' gestures from all possible perspectives. Recordings took place over two days. Some details about the 13 recorded conversations are available in Table~\ref{tab:recorded-conversations}\footnote{Interactions involving only blind speakers did not require the masked setting, which was aimed to let sighted speakers experience a sight impairment of some sort during in-presence communication.}.

\begin{table}
\caption{The number of recording sessions  in each considered condition (i.e., M=masked conversational situation, U=unmasked conversational situation, S=same sight conditions, D=different sight conditions). The total amount of recorded speech is also reported for each condition (in minutes).}
\begin{tabular}{c|llll}
\toprule
     &  M & U & &{\color{gray}mins}\\ 
     \midrule
   S  & 3 & 4 & 7 &{\color{gray}225.44}\\
   D & 3 & 3  & 6 & {\color{gray}202.71}\\
   \bottomrule
   & 6 & 7 \\
  {\color{gray}mins} & {\color{gray}198.1} & {\color{gray}225.4}
\end{tabular}
\label{tab:recorded-conversations}
\end{table}

\section{The \textit{Gest-IT} corpus schema}
\label{sec:schema}

The other decisions that we had to make from the very beginning concerned the repository, the archiving protocol, and the standards for transcribing the three layers we aim to represent (orthographic, prosodic and gestural). The next Sections are devoted to discuss these aspects in detail.

\subsection{Data repository}

Resource building is a team enterprise, performed asynchronously by a number of different people (i.e., PIs, interns, technicians etc.), often with different levels of technical expertise and background knowledge about the genesis of the data. Our project is no exception.

Therefore, in order to ensure data consistency and maintenance, a specific workflow has been put in place. More specifically, a central \texttt{git} repository\footnote{\url{https://github.com/LaboratorioSperimentale/Gest-IT}} keeps track of the status of the resource. 
The \texttt{main} branch contains the last, released version of the corpus while the \texttt{dev} branch is used for development in between releases (versions are numbered according to semantic versioning standards).

Each participant and each conversation is defined through a \texttt{.yaml} file (Appendix~\ref{app:metadata}), allowing for a number of CI/CD practices to be put in place: each time a new conversation description file is pushed to the repository, for instance, a table summarizing the full status of the resource is generated. Similarly, automatic checks are performed each time a transcription is updated to ensure the consistency of the overall resource: for instance, a script makes sure that names of layers in the transcription correspond to participants, that jeffersonian notation (see Section~\ref{sec:ortoproso}) is well formed, etc.

Data pertaining to each conversation is constituted by a set of digital objects, that represent different layers of information attached to the same recording. These include: (i) three video tracks and one audio track; (ii) a verbal transcription layer, which was initially automatically created with the \texttt{whisper} ASR toolkit~\cite{Radford2023} and then revised at the ortographic and prosodic level (Section~\ref{sec:ortoproso}); (iii) gesture transcription, starting from video sources (Section~\ref{sec:gesture}); (iv) UD annotation layers.

Transcriptions are maintained in CoNLL-U format\footnote{\url{https://universaldependencies.org/ext-format.html}}, with specific MISC features for the gesture component.
This will allow, in the future, to enrich the resource with additional annotation layers.

\subsection{Verbal language transcription}
\label{sec:ortoproso}

As regards verbal communication, we decided to adopt the standards of the KIParla corpus~\cite{mauri2019kiparla}, a corpus of spoken Italian that allows full access to audio files and transcriptions of roughly 153 hours of spontaneous speech~\footnote{The KIParla corpus is an incremental and modular resource, therefore this count refers to the three modules KIP, ParlaTO and KIPasti, which are online at the moment (as of July 2024). As soon as new modules are published, the global dimension of the resource will increase.}.

Once the recordings were acquired, the transcription process began. In accordance with the KIParla protocol, it was agreed to use the ELAN software~\cite{sloetjes2008annotation}, which allows for time alignment of videos, audio files and transcriptions.
In practice, the speech was segmented into transcription units identified on a perceptual basis, especially by reference to prosodic unit boundaries.
The transcription process involved two steps: 
\begin{itemize}
    \item \textit{orthographic transcription}, which included anonimization, turn assignment, and nonverbal behaviours. Whenever the annotator didn't understand, they could either choose `xxx' or type their hypothesis in parentheses; 
    \item \textit{prosodic transcription}, following a simplification of the Jefferson system~\cite{jefferson2004glossary}, widely shared by the scientific community~\cite{slembrouck2007transcription}. The employed conventions~\cite{ballare2022italiano} are reported in Table~\ref{tab:jefferson}.
\end{itemize}

\begin{table*}[]
    \centering
    \caption{Jefferson prosodic conventions}
    \begin{tabular}{llll}
    \toprule
Symbol & Description & Symbol & Description \\
\midrule
. & Descending intonation & ? & Rising intonation \\
(.) & Short pause & cia- & Interrupted word \\
\textgreater{}ciao\textless{} & Faster pronunciation & = & Prosodically bound units \\
, & Weakly rising intonation & \textless{}ciao\textgreater{} & Slower pronunciation \\
°ciao° & Lower volume & : & Prolonged sound \\
{[}ciao{]} & Overlap between speakers & CIAO & Louder volume \\
\bottomrule
\end{tabular}
\label{tab:jefferson}
\end{table*}

Transcriptions are thus available in two formats: they can be read as simple orthographic texts, or they can be read as enriched texts with prosodic and interactional information (such as overlaps, speed alterations, ascending or descending intonation, pauses, etc., as in example below). In both cases, it is possible to directly relate the transcription unit to the audiovisual unit. A further revision step will be done once the corpus will be fully transcribed, in order to make sure that notation is consistent throughout the resource.

\begin{exe}
\ex 
\gll S001 l'ultimo che ho fatt:o allora sono stata a siviglia:, p[er natale]\\
speak-id - - - Prolonged - - - - Prolonged+Ascending overlap overlap \\
\glt `my last one well I was in Seville for Christmas'
\ex 
\gll B001 [che io ador]o che [io adoro]\\
speak-id overlap overlap overlap - overlap overlap\\
\glt `(Seville) which I love'
\ex
\gll S001 [bellissima po]i a [natale è stato mag]ico\\
speak-id overlap overlap - overlap overlap overlap\\
\glt `wonderful Christmas was magic'
\ex 
\gll B001 [siviglia meravigliosa] \\
speak-id overlap overlap \\
\glt `wonderful Seville'
\end{exe}

\subsection{Gesture transcription}
\label{sec:gesture}

In order to provide also a transcription of gestures, as objective and interpretation-independent as possible, we decided to employ Typannot.

Typannot is a typographic system for the representation of sign languages, a project in development since 2013 by the \textit{Gestual Script} research group, composed of linguists, graphic designers, typographers, and computer scientists.  Its articulatory description of the body, independent of the language studied, allows it to be adapted to the study of gestures as well. Typannot proposes to analyze gestures and signs as realizations of the body and not just the hands: to facilitate analysis, the body is divided into different Articulatory Systems (AS), covering every body part from the hands to the feet and includes a description of facial expressions. For the purpose of the \textit{Gest-IT} project, only three will be considered:
\begin{description} 
    \item[Finger (F):] the dynamics of the fingers of the hand (thumb, index, middle, ring, and little finger). Furthermore, the distinction between the fingers of the right hand and those of the left hand will be considered and referred to respectively as RH and LH;
    \item[UpperLimb (UL):] the dynamics of the upper limbs (arm, forearm, hand);
    \item[UpperBody (UB):] the dynamics of the segments that make up torso (hip, spine and shoulder), neck and head.
\end{description}

In this system, the sign's form is seen as a set of articulatory body information (we extend this view to gestures). Currently, the generic characters that make up the graphic inventory of Typannot are used to describe the dynamics of all body segments.

\subsection{Towards a unified CoNLL-U corpus}
\label{sec:first}
The resulting corpus is composed of verbal-prosodic units and gestural units, with information about their overlaps\footnote{At the moment of writing, 1 minute of pilot transcription has been produced.}.
Each unit is described by the metadata listed in Table~\ref{tab:metadata}. In case of non verbal units, the text is filled with a placeholder token (\texttt{EMPTY}) and relevant information is contained in the MISC column, where the following features are introduced, \texttt{meta} for para-verbal information (such as laughs, coughs...) and \texttt{gesture} for Typannot codes (see Appendix~\ref{app:transcription}).

\begin{table}[]
    \centering
    \scriptsize{
    \begin{tabular}{p{0.3\columnwidth}|p{0.6\columnwidth}}
       sent\_id & unique identifier for transcritpion unit  \\
       text & space-separated sequence of token forms \\
       conversation\_id & unique identifier of conversatin \\
       speaker\_id & unique identifier of conversation participant \\
       duration &  duration of segment \\
       overlaps  & space-separated list of other sent\_ids\\
       text\_jefferson & for speech units, original prosodic/jeffersonian transcription \\
       type & for gestural units, identifier of the articulator \\
    \end{tabular}
    }
    \caption{Metadata describing each transcription (tu-xxx) or gestural unit (gu-xxx) in our corpus. \texttt{sent\_id} and \texttt{text} are derived from the UD format, while others are introduced for the purpose of this resource.}
    \label{tab:metadata}
\end{table}

\section{Future steps}
\label{sec:future}

The aim of this paper is to share with the scientific community the protocol developed to build a multimodal resource for the Italian language in terms of data collection (design, ethic issues, practicalities); data management and curation; data transcription, annotation and analysis. In doing so, we contribute to the debate on multimodal resource building, which is still lacking an established standard. In particular, our contribution in this respect is twofold.

Firstly, our study suggests to adopt a three-layer transcription where the three layers (i.e., the orthographic transcription, the prosodic/interactional transcription, and the gestural transcription) align to each other, by using ELAN as a tool for transcribing and CoNLL-X as an interoperable output format. This has the advantage of grounding gestures as an integrated semiotic source within verbal conversation and ultimately allows to unveil gesture-speech regularities.

Secondly, we propose an innovative approach for the annotation of gesture data. By relying on common practices in the field of sign languages, we suggest that gesture transcription should follow the same rationale of phonetic transcription, with a method that describes `objective' aspects that characterize the `form' of the gesture, thus allowing for an interpretation-independent annotation. 

Clearly, the project is still at a very preliminary stage. Next steps will include the complete orthographic, prosodic and gesture transcription of the recordings; a thorough revision and pseudoanymization.

\begin{acknowledgments}
The \textit{Gest-IT} corpus was built as an internship project at the \textit{Experimental Lab}\footnote{\url{https://site.unibo.it/laboratorio-sperimentale/}} of the Department of Modern Languages, Literatures, and Cultures (LILEC) of the University of Bologna. We would like to thank the \textit{Gestual Script} team (Ecole Supérieure d'Art et Design, ESAD, Amiens) for providing us with the Typannot system, and the \textit{Istituto dei ciechi Francesco Cavazza}\footnote{\url{https://www.cavazza.it/}} (Bologna) for helping us with the recruiting of visually impaired participants. A special acknowledgment to the \textit{Centro Studi sul Seicento e Settecento Spagnolo} (CSSS) of the LILEC Department for letting us record our pilot videos in their studio.
\end{acknowledgments}

\bibliography{sample-ceur}

\begin{thebibliography}{40}
\expandafter\ifx\csname natexlab\endcsname\relax\def\natexlab#1{#1}\fi
\providecommand{\url}[1]{\texttt{#1}}
\providecommand{\href}[2]{#2}
\providecommand{\path}[1]{#1}
\providecommand{\DOIprefix}{doi:}
\providecommand{\ArXivprefix}{arXiv:}
\providecommand{\URLprefix}{URL: }
\providecommand{\Pubmedprefix}{pmid:}
\providecommand{\doi}[1]{\href{http://dx.doi.org/#1}{\path{#1}}}
\providecommand{\Pubmed}[1]{\href{pmid:#1}{\path{#1}}}
\providecommand{\bibinfo}[2]{#2}
\ifx\xfnm\relax \def\xfnm[#1]{\unskip,\space#1}\fi
\bibitem[{Lüdeling and Kytö(2009)}]{ludeling2009corpus}
\bibinfo{author}{A.~Lüdeling}, \bibinfo{author}{M.~Kytö}, \bibinfo{title}{Corpus Linguistics: An International Handbook}, \bibinfo{publisher}{De Gruyter Mouton}, \bibinfo{year}{2009}. \URLprefix \url{https://www.degruyter.com/database/COGBIB/entry/cogbib.7917/html}.
\bibitem[{Bezemer and Jewitt(2018)}]{bezemer2018multimodality}
\bibinfo{author}{J.~Bezemer}, \bibinfo{author}{C.~Jewitt},
\newblock \bibinfo{title}{Multimodality: A guide for linguists},
\newblock \bibinfo{journal}{Research methods in linguistics} \bibinfo{volume}{28} (\bibinfo{year}{2018}).
\bibitem[{Abner et~al.(2015)Abner, Cooperrider, and Goldin-Meadow}]{abner2015gesture}
\bibinfo{author}{N.~Abner}, \bibinfo{author}{K.~Cooperrider}, \bibinfo{author}{S.~Goldin-Meadow},
\newblock \bibinfo{title}{Gesture for linguists: A handy primer},
\newblock \bibinfo{journal}{Language and Linguistics Compass} \bibinfo{volume}{9} (\bibinfo{year}{2015}) \bibinfo{pages}{437--451}. \DOIprefix\doi{10.1111/lnc3.12168}.
\bibitem[{Abuczki and Ghazaleh(2013)}]{abuczki2013overview}
\bibinfo{author}{{\'A}.~Abuczki}, \bibinfo{author}{E.~B. Ghazaleh},
\newblock \bibinfo{title}{An overview of multimodal corpora, annotation tools and schemes},
\newblock \bibinfo{journal}{Argumentum} \bibinfo{volume}{9} (\bibinfo{year}{2013}) \bibinfo{pages}{86--98}.
\bibitem[{Foster and Oberlander(2007)}]{foster2007corpus}
\bibinfo{author}{M.~E. Foster}, \bibinfo{author}{J.~Oberlander},
\newblock \bibinfo{title}{Corpus-based generation of head and eyebrow motion for an embodied conversational agent},
\newblock \bibinfo{journal}{Language Resources and Evaluation} \bibinfo{volume}{41} (\bibinfo{year}{2007}). \DOIprefix\doi{10.1007/s10579-007-9055-3}.
\bibitem[{Bressem et~al.(2013)Bressem, Ladewig, and Müller}]{Bressem2013}
\bibinfo{author}{J.~Bressem}, \bibinfo{author}{S.~H. Ladewig}, \bibinfo{author}{C.~Müller}, \bibinfo{title}{71. Linguistic Annotation System for Gestures}, \bibinfo{publisher}{De Gruyter Mouton}, \bibinfo{address}{Berlin, Boston}, \bibinfo{year}{2013}, pp. \bibinfo{pages}{1098--1124}. \URLprefix \url{https://doi.org/10.1515/9783110261318.1098}. \DOIprefix\doi{doi:10.1515/9783110261318.1098}.
\bibitem[{Bianchini(2024)}]{bianchini:hal-04602726}
\bibinfo{author}{C.~S. Bianchini}, \bibinfo{title}{{(D){\'e}crire les Langues des Signes: une approche grapholinguistique aux Langues des Signes}}, number~\bibinfo{number}{8} in \bibinfo{series}{Grapholinguistics and its Applications}, \bibinfo{publisher}{{Fluxus Editions}}, \bibinfo{year}{2024}. \URLprefix \url{https://hal.science/hal-04602726}. \DOIprefix\doi{10.36824/2024-bianchini}, \bibinfo{note}{iSSN 2681-8566 \& eISSN 2534-5192; EAN 9782487055025; CrossRef 1612798840}.
\bibitem[{Freigang et~al.(2014)Freigang, Priesters, Nishio, and Bergmann}]{freigang2014your}
\bibinfo{author}{F.~Freigang}, \bibinfo{author}{M.~A. Priesters}, \bibinfo{author}{R.~Nishio}, \bibinfo{author}{K.~Bergmann},
\newblock \bibinfo{title}{Your data at the center of attention: A metadata session profile for multimodal corpora},
\newblock in: \bibinfo{booktitle}{Proceedings of the CLARIN Annual Conference}, volume \bibinfo{volume}{2014}, \bibinfo{year}{2014}.
\bibitem[{L{\"u}cking et~al.(2010)L{\"u}cking, Bergmann, Hahn, Kopp, and Rieser}]{lucking2010bielefeld}
\bibinfo{author}{A.~L{\"u}cking}, \bibinfo{author}{K.~Bergmann}, \bibinfo{author}{F.~Hahn}, \bibinfo{author}{S.~Kopp}, \bibinfo{author}{H.~Rieser},
\newblock \bibinfo{title}{The bielefeld speech and gesture alignment corpus (saga)},
\newblock in: \bibinfo{booktitle}{LREC 2010 workshop: Multimodal corpora--advances in capturing, coding and analyzing multimodality}, \bibinfo{year}{2010}.
\bibitem[{Du~Bois and Troiani(2023)}]{dubois2023typology}
\bibinfo{author}{J.~Du~Bois}, \bibinfo{author}{G.~Troiani}, \bibinfo{title}{Typology and its data: functional monoculture or structural diversity?}, \bibinfo{howpublished}{presented at Naturally occurring data in and beyond linguistic typology}, \bibinfo{year}{2023}.
\bibitem[{Troiani(2023)}]{troiani2023representing}
\bibinfo{author}{G.~Troiani}, \bibinfo{title}{Representing a language in use: corpus construction, prosody, and grammar in Kazakh}, Ph.D. thesis, UC Santa Barbara, \bibinfo{year}{2023}.
\bibitem[{Bertrand et~al.(2008)Bertrand, Blache, Espesser, Ferr{\'e}, Meunier, Priego-Valverde, and Rauzy}]{bertrand2008cid}
\bibinfo{author}{R.~Bertrand}, \bibinfo{author}{P.~Blache}, \bibinfo{author}{R.~Espesser}, \bibinfo{author}{G.~Ferr{\'e}}, \bibinfo{author}{C.~Meunier}, \bibinfo{author}{B.~Priego-Valverde}, \bibinfo{author}{S.~Rauzy},
\newblock \bibinfo{title}{Le cid-corpus of interactional data-annotation et exploitation multimodale de parole conversationnelle},
\newblock \bibinfo{journal}{Revue TAL: traitement automatique des langues} \bibinfo{volume}{49} (\bibinfo{year}{2008}) \bibinfo{pages}{pp--105}.
\bibitem[{Knight et~al.(2009)Knight, Adolphs, Tennent, and Carter}]{knight2009nottingham}
\bibinfo{author}{D.~Knight}, \bibinfo{author}{S.~Adolphs}, \bibinfo{author}{P.~Tennent}, \bibinfo{author}{R.~Carter},
\newblock \bibinfo{title}{The nottingham multi-modal corpus: A demonstration},
\newblock in: \bibinfo{booktitle}{Programme of the Workshop on Multimodal Corpora}, \bibinfo{year}{2009}, p.~\bibinfo{pages}{64}.
\bibitem[{Trotta et~al.(2020)Trotta, Palmero~Aprosio, Tonelli, and Elia}]{trotta2020adding}
\bibinfo{author}{D.~Trotta}, \bibinfo{author}{A.~Palmero~Aprosio}, \bibinfo{author}{S.~Tonelli}, \bibinfo{author}{A.~Elia},
\newblock \bibinfo{title}{Adding gesture, posture and facial displays to the polimodal corpus of political interviews},
\newblock in: \bibinfo{booktitle}{Proceedings of the 12th Language Resources and Evaluation Conference (LREC 2020)}, \bibinfo{organization}{European Language Resources Association}, \bibinfo{year}{2020}, pp. \bibinfo{pages}{4320--4326}.
\bibitem[{Allwood(2001)}]{allwood2001capturing}
\bibinfo{author}{J.~Allwood},
\newblock \bibinfo{title}{Capturing differences between social activities in spoken language},
\newblock \bibinfo{journal}{Pragmatics and Beyond New Series}  (\bibinfo{year}{2001}) \bibinfo{pages}{301--320}.
\bibitem[{Allwood et~al.(2007)Allwood, Cerrato, Jokinen, Navarretta, and Paggio}]{allwood2007mumin}
\bibinfo{author}{J.~Allwood}, \bibinfo{author}{L.~Cerrato}, \bibinfo{author}{K.~Jokinen}, \bibinfo{author}{C.~Navarretta}, \bibinfo{author}{P.~Paggio},
\newblock \bibinfo{title}{The mumin coding scheme for the annotation of feedback, turn management and sequencing phenomena},
\newblock \bibinfo{journal}{Language Resources and Evaluation} \bibinfo{volume}{41} (\bibinfo{year}{2007}) \bibinfo{pages}{273--287}.
\bibitem[{P{\'a}pay et~al.(2011)P{\'a}pay, Szeghalmy, and Szekr{\'e}nyes}]{papay2011hucomtech}
\bibinfo{author}{K.~P{\'a}pay}, \bibinfo{author}{S.~Szeghalmy}, \bibinfo{author}{I.~Szekr{\'e}nyes},
\newblock \bibinfo{title}{Hucomtech multimodal corpus annotation},
\newblock \bibinfo{journal}{Argumentum} \bibinfo{volume}{7} (\bibinfo{year}{2011}) \bibinfo{pages}{330--347}.
\bibitem[{Schiel et~al.(2002)Schiel, Steininger, and T{\"u}rk}]{schiel2002smartkom}
\bibinfo{author}{F.~Schiel}, \bibinfo{author}{S.~Steininger}, \bibinfo{author}{U.~T{\"u}rk},
\newblock \bibinfo{title}{The smartkom multimodal corpus at bas.},
\newblock in: \bibinfo{booktitle}{LREC}, \bibinfo{organization}{Citeseer}, \bibinfo{year}{2002}.
\bibitem[{McNeill(2013)}]{mcneill2013gesture}
\bibinfo{author}{D.~McNeill}, \bibinfo{title}{Gesture and Thought}, \bibinfo{publisher}{University of Chicago Press}, \bibinfo{year}{2013}. \DOIprefix\doi{10.7208/chicago/9780226514642.001.0001}.
\bibitem[{Mlakar and Rojc(2012)}]{mlakar2012capturing}
\bibinfo{author}{I.~Mlakar}, \bibinfo{author}{M.~Rojc},
\newblock \bibinfo{title}{Capturing form of non-verbal conversational behavior for recreation on synthetic conversational agent eva},
\newblock \bibinfo{journal}{WSEAS Trans. Comput.[Print ed.]} \bibinfo{volume}{11} (\bibinfo{year}{2012}) \bibinfo{pages}{218--226}.
\bibitem[{Mlakar et~al.(2019)Mlakar, Verdonik, Majheni{\v{c}}, and Rojc}]{mlakar2019towards}
\bibinfo{author}{I.~Mlakar}, \bibinfo{author}{D.~Verdonik}, \bibinfo{author}{S.~Majheni{\v{c}}}, \bibinfo{author}{M.~Rojc},
\newblock \bibinfo{title}{Towards pragmatic understanding of conversational intent: A multimodal annotation approach to multiparty informal interaction--the eva corpus},
\newblock in: \bibinfo{booktitle}{Statistical Language and Speech Processing: 7th International Conference, SLSP 2019, Ljubljana, Slovenia, October 14--16, 2019, Proceedings 7}, \bibinfo{organization}{Springer}, \bibinfo{year}{2019}, pp. \bibinfo{pages}{19--30}.
\bibitem[{Fi{\v{s}}er and Lenardi{\v{c}}(2020)}]{fivser2020overview}
\bibinfo{author}{D.~Fi{\v{s}}er}, \bibinfo{author}{J.~Lenardi{\v{c}}},
\newblock \bibinfo{title}{Overview of multimodal corpora in the clarin}  (\bibinfo{year}{2020}).
\bibitem[{MacWhinney(1998)}]{macwhinney1998computational}
\bibinfo{author}{B.~MacWhinney},
\newblock \bibinfo{title}{Computational transcript analysis and language disorders},
\newblock in: \bibinfo{booktitle}{Handbook of Neurolinguistics}, \bibinfo{publisher}{Elsevier}, \bibinfo{year}{1998}, pp. \bibinfo{pages}{599--616}.
\bibitem[{Lo~Re(2021)}]{lo2021prosody}
\bibinfo{author}{L.~Lo~Re},
\newblock \bibinfo{title}{Prosody and gestures to modelling multimodal interaction: Constructing an italian pilot corpus},
\newblock \bibinfo{journal}{IJCoL. Italian Journal of Computational Linguistics} \bibinfo{volume}{7} (\bibinfo{year}{2021}) \bibinfo{pages}{33--44}.
\bibitem[{Kendon et~al.(1980)}]{kendon1980gesticulation}
\bibinfo{author}{A.~Kendon}, et~al.,
\newblock \bibinfo{title}{Gesticulation and speech: Two aspects of the process of utterance},
\newblock \bibinfo{journal}{The relationship of verbal and nonverbal communication} \bibinfo{volume}{25} (\bibinfo{year}{1980}) \bibinfo{pages}{207--227}.
\bibitem[{Lo~Re(2022)}]{lo2022corpus}
\bibinfo{author}{L.~Lo~Re}, \bibinfo{title}{Corpus multimodale dell’italiano parlato: basi metodologiche per la creazione di un prototipo}, Ph.D. thesis, University of Firenze, \bibinfo{year}{2022}.
\bibitem[{Ackerley and Coccetta(2007)}]{AckerleyCoccetta2007}
\bibinfo{author}{K.~Ackerley}, \bibinfo{author}{F.~Coccetta},
\newblock \bibinfo{title}{Enriching language learning through a multimedia corpus},
\newblock \bibinfo{journal}{ReCALL} \bibinfo{volume}{19} (\bibinfo{year}{2007}) \bibinfo{pages}{351–370}. \DOIprefix\doi{10.1017/S0958344007000730}.
\bibitem[{Coccetta et~al.(2011)}]{coccetta2011multimodal}
\bibinfo{author}{F.~Coccetta}, et~al.,
\newblock \bibinfo{title}{Multimodal functional-notional concordancing},
\newblock \bibinfo{journal}{New Trends in Corpora and Language Learning. London: Continuum}  (\bibinfo{year}{2011}) \bibinfo{pages}{121--138}.
\bibitem[{Rohrer et~al.(2020)Rohrer, Vil{\`a}-Gim{\'e}nez, Florit-Pons, Esteve-Gibert, Ren, Shattuck-Hufnagel, and Prieto}]{rohrer2020multimodal}
\bibinfo{author}{P.~L. Rohrer}, \bibinfo{author}{I.~Vil{\`a}-Gim{\'e}nez}, \bibinfo{author}{J.~Florit-Pons}, \bibinfo{author}{N.~Esteve-Gibert}, \bibinfo{author}{A.~Ren}, \bibinfo{author}{S.~Shattuck-Hufnagel}, \bibinfo{author}{P.~Prieto},
\newblock \bibinfo{title}{The multimodal multidimensional (m3d) labelling scheme for the annotation of audiovisual corpora},
\newblock \bibinfo{journal}{Gesture and Speech in Interaction (GESPIN)}  (\bibinfo{year}{2020}).
\bibitem[{Rohrer et~al.(2023)Rohrer, Delais-Roussarie, and Prieto}]{rohrer2023visualizing}
\bibinfo{author}{P.~L. Rohrer}, \bibinfo{author}{E.~Delais-Roussarie}, \bibinfo{author}{P.~Prieto},
\newblock \bibinfo{title}{Visualizing prosodic structure: Manual gestures as highlighters of prosodic heads and edges in english academic discourses},
\newblock \bibinfo{journal}{Lingua} \bibinfo{volume}{293} (\bibinfo{year}{2023}) \bibinfo{pages}{103583}.
\bibitem[{Chevrefils et~al.(2021)Chevrefils, Danet, Doan, Thomas, R{\'e}bulard, Adrien, Dauphin, and Bianchini}]{chevrefils2021body}
\bibinfo{author}{L.~Chevrefils}, \bibinfo{author}{C.~Danet}, \bibinfo{author}{P.~Doan}, \bibinfo{author}{C.~Thomas}, \bibinfo{author}{M.~R{\'e}bulard}, \bibinfo{author}{C.~Adrien}, \bibinfo{author}{J.-F. Dauphin}, \bibinfo{author}{C.~S. Bianchini},
\newblock \bibinfo{title}{The body between meaning and form: kinesiological analysis and typographical representation of movement in sign languages},
\newblock \bibinfo{journal}{Languages and Modalities} \bibinfo{volume}{1} (\bibinfo{year}{2021}) \bibinfo{pages}{49--63}.
\bibitem[{Boutet et~al.(2018)Boutet, Doan, Danet, Bianchini, Goguely, Contesse, and R{\'e}bulard}]{boutet2018systemes}
\bibinfo{author}{D.~Boutet}, \bibinfo{author}{P.~Doan}, \bibinfo{author}{C.~Danet}, \bibinfo{author}{C.~S. Bianchini}, \bibinfo{author}{T.~Goguely}, \bibinfo{author}{A.~Contesse}, \bibinfo{author}{M.~R{\'e}bulard},
\newblock \bibinfo{title}{Syst{\`e}mes graph{\'e}matiques et {\'e}critures des langues sign{\'e}es},
\newblock \bibinfo{journal}{Signata. Annales des s{\'e}miotiques/Annals of Semiotics}  (\bibinfo{year}{2018}) \bibinfo{pages}{391--426}.
\bibitem[{Alibali et~al.(2001)Alibali, Heath, and Myers}]{Alibali2001}
\bibinfo{author}{M.~W. Alibali}, \bibinfo{author}{D.~C. Heath}, \bibinfo{author}{H.~J. Myers},
\newblock \bibinfo{title}{Effects of visibility between speaker and listener on gesture production: Some gestures are meant to be seen},
\newblock \bibinfo{journal}{Journal of Memory and Language} \bibinfo{volume}{44} (\bibinfo{year}{2001}) \bibinfo{pages}{169--188}. \DOIprefix\doi{10.1006/JMLA.2000.2752}.
\bibitem[{Iverson and Goldin-Meadow(1998)}]{Iverson1998}
\bibinfo{author}{J.~M. Iverson}, \bibinfo{author}{S.~Goldin-Meadow},
\newblock \bibinfo{title}{Why people gesture when they speak},
\newblock \bibinfo{journal}{Nature 1998 396:6708} \bibinfo{volume}{396} (\bibinfo{year}{1998}) \bibinfo{pages}{228--228}. \URLprefix \url{https://www.nature.com/articles/24300}. \DOIprefix\doi{10.1038/24300}.
\bibitem[{Radford et~al.(2023)Radford, Kim, Xu, Brockman, Mcleavey, and Sutskever}]{Radford2023}
\bibinfo{author}{A.~Radford}, \bibinfo{author}{J.~W. Kim}, \bibinfo{author}{T.~Xu}, \bibinfo{author}{G.~Brockman}, \bibinfo{author}{C.~Mcleavey}, \bibinfo{author}{I.~Sutskever},
\newblock \bibinfo{title}{Robust speech recognition via large-scale weak supervision},
\newblock \bibinfo{year}{2023}.
\bibitem[{Mauri et~al.(2019)Mauri, Ballar{\`e}, Goria, Cerruti, Suriano et~al.}]{mauri2019kiparla}
\bibinfo{author}{C.~Mauri}, \bibinfo{author}{S.~Ballar{\`e}}, \bibinfo{author}{E.~Goria}, \bibinfo{author}{M.~Cerruti}, \bibinfo{author}{F.~Suriano}, et~al.,
\newblock \bibinfo{title}{Kiparla corpus: a new resource for spoken italian},
\newblock in: \bibinfo{booktitle}{CEUR WORKSHOP PROCEEDINGS}, \bibinfo{organization}{SunSITE Central Europe}, \bibinfo{year}{2019}, pp. \bibinfo{pages}{1--7}.
\bibitem[{Sloetjes and Wittenburg(2008)}]{sloetjes2008annotation}
\bibinfo{author}{H.~Sloetjes}, \bibinfo{author}{P.~Wittenburg},
\newblock \bibinfo{title}{Annotation by category-elan and iso dcr},
\newblock in: \bibinfo{booktitle}{6th international Conference on Language Resources and Evaluation (LREC 2008)}, \bibinfo{year}{2008}.
\bibitem[{Jefferson et~al.(2004)}]{jefferson2004glossary}
\bibinfo{author}{G.~Jefferson}, et~al.,
\newblock \bibinfo{title}{Glossary of transcript symbols with an introduction},
\newblock \bibinfo{journal}{Conversation analysis}  (\bibinfo{year}{2004}) \bibinfo{pages}{13--31}.
\bibitem[{Slembrouck(2007)}]{slembrouck2007transcription}
\bibinfo{author}{S.~Slembrouck},
\newblock \bibinfo{title}{Transcription—the extended directions of data histories: a response to m. bucholtz's' variation in transcription'},
\newblock \bibinfo{journal}{Discourse Studies} \bibinfo{volume}{9} (\bibinfo{year}{2007}) \bibinfo{pages}{822--827}.
\bibitem[{Ballar{\`e} et~al.(2022)Ballar{\`e}, Goria, and Mauri}]{ballare2022italiano}
\bibinfo{author}{S.~Ballar{\`e}}, \bibinfo{author}{E.~Goria}, \bibinfo{author}{C.~Mauri}, \bibinfo{title}{Italiano parlato e variazione linguistica: Teoria e prassi nella costruzione del corpus KIParla}, \bibinfo{publisher}{P{\`a}tron editore}, \bibinfo{year}{2022}.

\end{thebibliography}

\appendix

\section{Prompts for conversation}
\label{app:prompts}
\begin{enumerate}
    \item Sai che a Bologna c'è questa storia dell'``umarel''? Sai cos'è un umarel?

    \item Sai quante lingue si insegnano al LILEC, il dipartimento di lingue dell'università di Bologna? Sapresti elencarle? 

    \item C'è una lingua che hai sempre voluto imparare? E una che invece proprio non ti ha mai incuriosito? 

    \item Secondo te quante sono le lingue parlate nel mondo? 

    \item Alcune espressioni sono veramente curiose: per esempio, hai mai pensato come mai la ``zuppa inglese'' si chiama così? 

    \item Parli un dialetto? Con chi lo parli? Quando lo parli? 

    \item Alcune espressioni sono veramente curiose: per esempio, hai mai pensato come mai si dice ``fumare come un turco''? 

    \item I tortellini bolognesi: ti piacciono o no? Ma perché costano così tanto? 

    \item Qual è un piatto della tua infanzia che ricordi sempre con piacere? 

    \item Quale piatto cucini più spesso? Come lo prepari? Che ingredienti usi? 

    \item In che zona di Bologna vivi? Ti piace? Perché? 

    \item Secondo te, possono esistere lingue con massimo due parole per indicare i colori?  

    \item Che differenza c'è tra un dialetto e una lingua? 

    \item Ma perché si dice “chi va a Roma perde la poltrona”?  

    \item Credi che Bologna sia una città sicura dove vivere? Quali sono i suoi pro e i suoi contro? 

    \item In Italia il dialetto è un vero e proprio simbolo identitario. E tu che rapporto hai con il dialetto? Lo parli spesso? E con chi? 

    \item Qual è viaggio ti ha lasciato il ricordo più bello?  

    \item Chi è il tuo/la tua cantante preferito/a? Hai mai avuto modo di assistere a un suo concerto?  

    \item Secondo te esistono lingue più facili o più difficili da imparare, che per te suonano meglio o peggio? Quali e perché? 

    \item Hai qualche sogno o obiettivo che stai cercando di realizzare? 

    \item Se potessi vivere in un'altra città, quale sarebbe e perché?  

    \item C'è una lingua che avresti sempre voluto imparare, ma non hai mai studiato? Cosa ti attrae di questa lingua? 

    \item Credi che l'apprendimento di una nuova lingua possa influenzare il modo in cui vedi il mondo? In che modo?  
    \item Hai mai avuto difficoltà a comprendere gli accenti regionali o le varietà linguistiche?  

    \item Qual è il tuo modo preferito per rilassarti dopo una lunga giornata? 

    \item Se dovessi spiegare un modo di dire italiano a qualcuno che non lo conosce, quale sceglieresti e come lo spiegheresti? 

    \item Qual è il modo di dire italiano che trovi particolarmente divertente o curioso? 

    \item Cosa pensi del dibattito sull'influenza dell'inglese sull'italiano contemporaneo? È una minaccia o un arricchimento? 

    \item La lingua italiana è considerata una delle più musicali al mondo. Secondo te è vero? Quali sono, secondo te, altre lingue particolarmente musicali? E quali invece non lo sono affatto? 

    \item Hai mai avuto l'occasione di assaggiare la cucina tipica di un'altra nazione? Quale piatto ti è piaciuto in particolar modo e quale invece non ti ha convinto a pieno? 

\end{enumerate}

\section{Metadata schemata}
\label{app:metadata}

For both participants (see Subsection~\ref{subs:participants}) and conversations Subsection~\ref{subs:conversations}), metadata is collected and maintained in \texttt{.yaml} files, with the following formats

\subsection{Participants metadata}
\label{subs:participants}

\begin{lstlisting}

    Code: # 4-char string composed by either S (Sighted) or B (Blind) and an integer padded with 0s

    Gender: # either F (Female) or M (Male)

    Age: # age range of the participant expressed as 5-years bins (0-5, 6-10, 11-20,...)

    Region: # 1 of the 20 italian regions (typing conventions provided)

    First language: # upper cased iso-693-3 code of mother tongue

    Education level: # one value in (Primaria, Medie inferiori, Medie superiori, Laurea, PhD)

    Profession: # istat-derived category for profession (list provided)

    Notes on sight-related disabilities: # any relevant annotation on sight-related conditions declared by the participant
\end{lstlisting}

\subsection{Conversation metadata}
\label{subs:conversations}

\begin{lstlisting}
    Code: # 11-char string composed by [D|S] (same-condition or different-condition participant) + [M|U] (masked or unmasked conversation) + [L|S] (code associated to room where the conversation was recorded) + [DDMMhhmm]

    Participants:
      - [participant_code_1] # code of participant sitting on left side
      - [participant_code_2] # code of participant sitting on right side

    Facing: # M (Masked) or U (unmasked) depending on type of conversation

    Data:
      - Video:
        - Left: path/to/left/camera/recording
        - Centre: path/to/central/camera/recording
        - Right: path/to/right/camera/recording
      - Audio: path/to/audio/file
      - Transcription:
        - Automatic: path/to/automatic/transcription
        - Manually revised: path/to/manually/revised/transcription
        - Prosodic: path/to/prosodic/transcription
        - Gestual: path/to/gestual/transcription
\end{lstlisting}

\section{Integrated transcription in ELAN}
\label{app:transcription}

\begin{figure*}[h]
    \centering
    \includegraphics[width=0.8\textwidth]{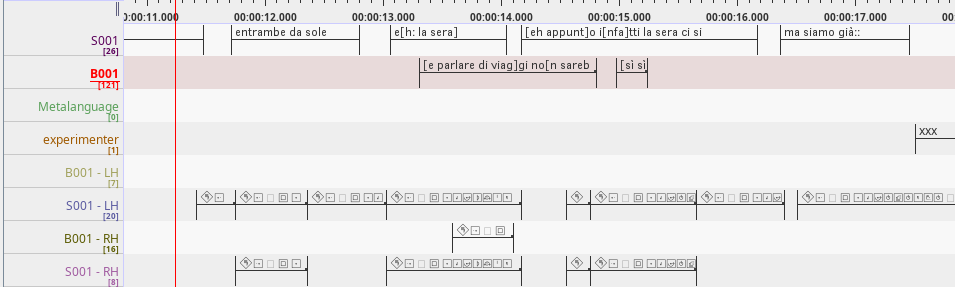}
    \caption{ELAN tiers showing verbal and gestural transcriptions.}
    \label{fig:elan-environment}
\end{figure*}

Figure~\ref{fig:elan-environment} shows an example of the collaborative ELAN environment where transcriptions are developed. The picture shows 8 tiers: the first two (\texttt{S001} and \texttt{B001}) refer to the verbal-prosodic transcription; an additional \texttt{experimenter} tier is used to take care of verbal productions of the experimenter, in case they occurr; the \textit{metalanguage} tier encodes non-verbal acts such laughs, noises etc.; the remaining tiers encode the Typannot-based transcriptions for finger articulators (\texttt{F:LH} and \texttt{F:RH} stand for \textit{left hand} and \textit{right hand} respectively).

The full CoNLL-U data can be consulted at \url{https://github.com/LaboratorioSperimentale/gest-IT/blob/dev/data/conll/DUC22051430.annotated.conll}, a small portion is reported in Figure~\ref{fig:conll}.

\begin{figure*}[t]
\caption{CoNLL-U extract}
\scriptsize{
\begin{lstlisting}
# sent_id = tu0005
# overlaps = gu0003 gu0004 gu0005 gu0006
# conversation = DUC22051430
# speaker_id = S001
# duration = 1.088
# text_jefferson = entrambe da sole
# text = entrambe da sole
1  entrambe  entrambi  PRON  _  Gender=Fem|Number=Plur|PronType=Ind  0  root  _  AlignBegin=11.704
2  da  da  ADP  _  _  3  case  _  _
3  sole  solo  ADJ  _  Gender=Fem|Number=Plur  1  nmod  _  AlignEnd=12.792

# sent_id = tu0006
# overlaps = tu0007 gu0007 gu0008 gu0009
# conversation = DUC22051430
# speaker_id = S001
# duration = 0.987
# text_jefferson = e[h: la sera]
# text = eh la sera
1  eh  eh  INTJ  _  _  3  discourse  _  AlignBegin=13.047|Overlap=B:tu0007|ProlongedSound=eh:
2  la  il  DET  _  Definite=Def|Gender=Fem|Number=Sing  3  det  _  Overlap=I
3  sera  sera  NOUN  _  Gender=Fem|Number=Sing  0  root  _  AlignEnd=14.034|Overlap=I

# sent_id = tu0007
# overlaps = tu0006 tu0008 gu0007 gu0008 gu0009 gu0010 gu0011 gu0012 gu0013
# conversation = DUC22051430
# speaker_id = B001
# duration = 1.500
# text_jefferson = [e parlare di viag]gi no[n sarebbe male]
# text = e parlare di viaggi non sarebbe male
1  e  e  CCONJ  _  _  7  cc  _  AlignBegin=13.3|Overlap=B:tu0006
2  parlare  parlare  VERB  _  VerbForm=Inf  7  csubj  _  Overlap=I
3  di  di  ADP  _  _  4  case  _  Overlap=I
4  viaggi  viaggio  NOUN  _  Gender=Masc|Number=Plur  2  obl  _  Overlap=I
5  non  non  ADV  _  _  7  advmod  _  Overlap=B:tu0008
6  sarebbe  essere  AUX  _  Mood=Cnd|Number=Sing|Person=3|Tense=Pres|VerbForm=Fin  7  cop  _  Overlap=I
7  male  male  ADV  _  _  0  root  _  AlignEnd=14.8|Overlap=I

# conversation = DUC22051430
# sent_id = gu0003
# overlaps = tu0004 tu0005
# speaker = S001
# duration = 0.330
# text = EMPTY
# type = F:LH
1  EMPTY  EMPTY  X  _  _  0  root  _  AlignBegin=11.410|AlignEnd=11.740|gesture='\ue5de\ue002 [ \uf197 \ue008\uf19f\ue5ea\ue5ef\ue5e8\ue5ef\uf1a0\ue5fe\ue5ee - \ue004\ue005\ue006\ue007\uf19f\ue5fe\ue5ee\ue5e8\ue5ef\uf1a0\ue5e7\ue5ef ] [ \uf198\ue001 ]'

# conversation = DUC22051430
# sent_id = gu0004
# overlaps = tu0005 gu0005
# speaker = S001
# duration = 0.610
# text = EMPTY
# type = F:LH
1  EMPTY  EMPTY  X  _  _  0  root  _  AlignBegin=11.740|AlignEnd=12.350|gesture='\ue5de\ue002 [ \uf197 \ue008\uf19f\ue5ff\ue5ee\ue5fb\ue5ee\uf1a0\ue5fe\ue5ee - \ue004\ue005\ue006\ue007\uf19f\ue5fe\ue5ee\ue5fb\ue5ee\uf1a0\ue5fd\ue5ee ] [ \uf198\ue001 ]'

# conversation = DUC22051430
# sent_id = gu0005
# overlaps = tu0005 gu0004
# speaker = S001
# duration = 0.610
# text = EMPTY
# type = F:RH
1  EMPTY  EMPTY  X  _  _  0  root  _  AlignBegin=11.740|AlignEnd=12.350|gesture='\ue5de\ue003 [ \uf197 \ue008\uf19f\ue5ff\ue5ee\ue5fb\ue5ee\uf1a0\ue5fe\ue5ee - \ue004\ue005\ue006\ue007\uf19f\ue5fe\ue5ee\ue5fb\ue5ee\uf1a0\ue5fd\ue5ee ] [ \uf198\ue001 ]'

# conversation = DUC22051430
# sent_id = gu0006
# overlaps = tu0005
# speaker = S001
# duration = 0.670
# text = EMPTY
# type = F:LH
1  EMPTY  EMPTY  X  _  _  0  root  _  AlignBegin=12.350|AlignEnd=13.020|gesture='\ue5de\ue002 [ \uf197 \ue008\uf19f\ue5ea\ue5ef\ue5fb\ue5ee\uf1a0\ue5fe\ue5ee - \ue004\uf19f\ue5e7\ue5ef\ue5fb\ue5ee\uf1a0\ue5fd\ue5ee - \ue005\ue006\ue007\uf19f\ue5fe\ue5ee\ue5fb\ue5ee\uf1a0\ue5fd\ue5ee ] [ \uf198 ( \ue00e\ue001 ) ( \ue00b \ue00c\ue005\ue006\ue007 - \uf196\ue5ee\ue001 ) ]'


\end{lstlisting}
}
\label{fig:conll}
\end{figure*}

\end{document}